\pdfoutput=1

\documentclass[11pt]{article}

\usepackage[]{EACL2023}

\usepackage{times}
\usepackage{latexsym}

\usepackage{amsmath}
\usepackage{graphicx}

\newcommand\scalemath[2]{\scalebox{#1}{\mbox{\ensuremath{\displaystyle #2}}}}

\usepackage[T1]{fontenc}

\usepackage[utf8]{inputenc}

\usepackage{microtype}

\usepackage{inconsolata}

\title{Leveraging Task Dependency and  Contrastive Learning for Case Outcome Classification on European Court of Human Rights Cases}

\author{ Santosh T.Y.S.S, \bf{Marcel Perez San Blas, Phillip Kemper, Matthias Grabmair} \\ School of Computation, Information, and Technology; \\
Technical University of Munich, Germany \\ \ { \texttt{\{santosh.tokala, marcel.perez, phillip.kemper, matthias.grabmair\}@tum.de} }}

\begin{document}
\maketitle
\begin{abstract}
We report on an experiment in case outcome classification on European Court of Human Rights cases where our model first learns to identify the convention articles allegedly violated by the state from case facts descriptions, and subsequently uses that information to classify whether the court finds a violation of those articles. We assess the dependency between these two tasks at the feature and outcome level. Furthermore, we leverage a hierarchical contrastive loss to pull together article-specific representations of cases at the higher level, leading to distinctive article clusters. The cases in each article cluster are further pulled closer based on their outcome, leading to sub-clusters of cases with similar outcomes. Our experiment results demonstrate that, given a static pre-trained encoder, our models produce a small but consistent improvement in classification performance over single-task and joint models without contrastive loss.
\end{abstract}

\section{Introduction}
\label{intro}
The NLP task of classifying case outcome information from a textual statement of case facts is generally referred to as Legal Judgement Prediction (LJP)(e.g., \citealt{aletras2016predicting, chalkidis2019neural}). It has been studied using corpora from different jurisdictions, such as the European Court of Human Rights (ECtHR) \cite{chalkidis2019neural, chalkidis2021paragraph, chalkidis2022lexglue, aletras2016predicting, medvedeva2020using, santosh2022deconfounding,santosh2023Zero},
Chinese Criminal Courts \cite{luo2017learning,zhong2018legal, yue2021neurjudge, zhong2020iteratively,yang2019legal},
 US Supreme Court \cite{katz2017general,kaufman2019improving}, Indian Supreme Court \cite{malik2021ildc,shaikh2020predicting},
the French court of Cassation \cite{csulea2017predicting},
Brazilian courts \cite{ bertalan2020predicting},
the Federal Supreme Court of Switzerland \cite{niklaus2021swiss}, 
the Turkish Constitutional court \cite{sert2021using,mumcuouglu2021natural},
 UK courts \cite{strickson2020legal}, German courts \cite{waltl2017predicting},
 the Philippine Supreme court \cite{virtucio2018predicting}, and the Thailand Supreme Court \cite{kowsrihawat2018predicting}. 

In this work, we experiment with classifying case outcomes in the ECtHR A and B benchmark tasks introduced by LexGLUE \cite{chalkidis2022lexglue}. Task B is to identify the set of articles of the European Convention of Human Rights (ECHR) that the claimant alleges to have been violated, while Task A's goal is to classify which of the convention’s articles has been deemed violated by the court. The input for both tasks is the case’s fact description that has been extracted from the published judgement document. It should be noted that, despite these tasks being typically referred to as instances of `legal judgement prediction', the fact statements are typically not finalized until the decision outcome is known, making the task effectively one of retrospective classification rather than prediction \cite{medvedeva2021automatic}. While this does lead to distracting and confounding phenomena (see our prior work in \citealt{santosh2022deconfounding}), the dataset remains a useful resource for the development of NLP models that analyze these fact statements for text patterns that correspond to specific convention articles as drafted by the court. Consequently, we speak of \textit{case outcome classification (COC)}. 

Positive instances for Task A are cases in which an article was deemed violated by the court. Negative instances, however, usually encompass not only cases in which that particular article was alleged and considered not violated, but also the cases in which the particular article was not alleged in the first place. Given that the conditional probability of a positive task A label (violation) given its task B label (allegation) can be very high in the LexGLUE dataset, we posit that models can fall into this pitfall of identifying dominant articles with high conditional violation probability, missing information specific to violation classification. Thus, we experiment with multi-task architectures to decouple these reasoning steps 
involved in Task A . This is similar to work in Chinese criminal cases judgement prediction carried out in \citealt{zhong2018legal, yang2019legal, xu2020distinguish,yue2021neurjudge} that coordinated multiple outcome variables (law articles, criminal charges, and penalty terms).

In the ECtHR context, most of the previous works  \cite{chalkidis2019neural,chalkidis2021paragraph, chalkidis2022lexglue, clavie2021unreasonable, santosh2022deconfounding,santosh2023Zero} employed independent models for allegation identification and violation classification. Concurrent work by \citealt{valvoda2022role} has recently explored multi-task joint models that learn Task A and Task B simultaneously. Their \emph{Claim-Outcome model} decomposes them into two independent classifiers with separate encoders for each allegation and violation classification given allegation information. Our work differs in three ways: (i) We learn the joint representation for both tasks through a shared encoder following our intuition that the shared representation is beneficial as both tasks require similar features at the lower level; (ii) We model dependency from allegation to violation classification at both the allegation outcome and feature levels. Conditioning violation classification on the allegation outcome leads to higher performance, but we find that passing along feature level information yields additional improvement. This way, the violation branch will focus on identifying  specific information required for determining violation than falling prey to only allegation level features. (iii) We also model correlations among different articles that tend to be concurrently alleged.

Inspired by recent advances in contrastive learning to learn effective representations \cite{khosla2020supervised}, we further devise a two-level hierarchical contrastive loss for COC. First, we strive to maximize the latent space distance between different article representations, which assists the model in learning distinct article-specific views of case facts. Second, we apply contrastive learning within each article latent space to form distinct sub-clusters of similar outcome cases. Unlike our two-level hierarchical contrastive learning, a single-level one has been explored for LJP in the concurrent work by \citealt{zhang2023contrastive}.

Our experiments demonstrate that, given a static pre-trained encoder, our models outperform  single-task and joint models without contrastive loss by a small but consistent margin, with larger improvements for sparse classes in classification performance.

\section{Method}
\begin{figure}[]
\centering
 \scalebox{.48}{
\includegraphics[width =\textwidth]{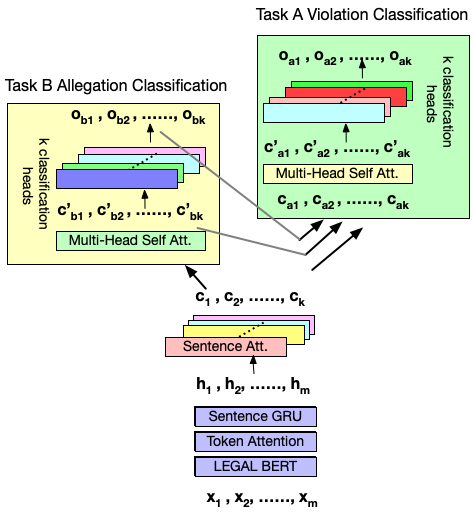}}
\caption{Our architecture capturing task dependency from allegation to violation classification branch.}
\label{arch}
\end{figure}

Our model takes as input the case fact description as token encodings $x = \{x_1, x_2, \ldots, x_m\}$ where $x_i = \{x_{i1}, x_{i2},\ldots,x_{in}\}$ and outputs the set of articles claimed to be alleged by the applicant (Task B) and set of articles deemed to be violated by the court (Task A) as multi-hot vectors. $x_i$  and $x_{ip}$ denote $i^\text{th}$ sentence and $p^\text{th}$ token of $i^\text{th}$ sentence of the case facts respectively. $m$ and $n$ denote the number of sentences and tokens in the $i^\text{th}$ fact sentence, respectively. Our model is built on top of a hierarchical attention network \cite{yang2016hierarchical}, which we chose to be able to process the very long fact description inputs. Our model contains three main components: (1) encoding layer, (2) Article-Specific Case representation and (3) Task Dependency learning. See Fig. \ref{arch} for an overview of our model. 

\noindent \textbf{Encoding layer:} We encode each sentence in the case facts description with LegalBERT \cite{chalkidis2020legal} to obtain the token level representations $\{z_{i1}, z_{i2},\ldots, z_{in}\}$. Sentence level representations are then computed using a token-level attention mechanism as follows: 
\begin{equation}
\scalemath{0.7}{
    u_{it} = \text{tanh}(W_w z_{it} + b_w ) ~~\& ~~ 
    \alpha_{it} = \frac{\text{exp}(u_{it}u_w)}{\sum_t \text{exp}(u_{it}u_w)}  ~~\& ~~ 
    f_i = \sum_{t=1}^n \alpha_{it}z_{it}}
\label{att}
\end{equation}
where $W_w$, $b_w$ and $u_w$ are trainable parameters. The sentence level representations $\{f_1, \ldots, f_m\}$ are passed through a GRU encoder to obtain context-aware representations of the case facts $h = \{h_1, h_2, \ldots, h_m\}$. 

\noindent \textbf{Article-Specific Case Representation:} We disentangle the input case facts into multiple article-view-representations   $ c = \{c_1, c_2,\ldots, c_k\}$ where $k$ is the total number of modeled convention articles, through aggregating context-aware sentence-level representations $h$ using $k$ sentence-level attention mechanisms similar to eq. \ref{att} for every article individually. Through this article-specific attention mechanism,  sentences relevant to  a specific article are emphasized and intended to aid in fine-grained reasoning of outcome prediction for every article. This is helpful especially for sparser classes and mitigates the tendency of models to focus on skewed dominant articles. This is distinct from previous works \cite{chalkidis2019neural,chalkidis2021paragraph,chalkidis2022lexglue, santosh2022deconfounding} which use a single vector representation for case facts.

\noindent \textbf{Task Dependency Learning:} To capture both the inter-article correlations for allegation, violation classification, and the inter-task dependency, we again use a two-step architecture. First, we apply a multi-head self-attention layer \cite{vaswani2017attention} to the obtained  article-specific representations of case $c$ to allow interactions among articles (e.g., some articles are typically alleged together while others usually occur in isolation) and obtain article-interaction-aware allegation feature representations $c^\prime_b =  \{c^\prime_{b1}, c^\prime_{b2},\ldots, c^\prime_{bk}\}$. These $c^\prime_b$ are passed through $k$ article-specific Task B classification layers to obtain the binary outcome $o_b$ corresponding to each article for allegation classification, which are then concatenated into a multi-hot vector. 

Then we utilize the task B allegation information to enhance task A violation classification to capture the task dependency. 
We concatenate the obtained article-interaction-aware allegation representations $c^\prime_b$ and allegation label probability logits $o^\prime_b$ \footnote{We use non-binarized response (i.e. probability logits)  as it avoids the information loss that can occur in the  probability space due to binarization.} 
with article-specific case representations $c$ to obtain the enhanced article-specific representation for Task A as $c_{ai} = [c_i,c^\prime_{bi},o^\prime_{bi}]$. We again employ multi-head self-attention mechanism on these enhanced article-aware representations $c_a$ for violation prediction to capture the correlations that exist between violation of different articles. Finally, the obtained representations are passed through $k$ article-specific Task A classification layers to obtain the binary outcome $o_a$ corresponding to each article's violation output. To be able to evaluate the effectiveness of our interaction architecture in a clean way, we freeze the LegalBERT encoder weights in all our experiments.

\subsection{Hierarchical Contrastive Loss} 
Contrastive learning has recently gained attention as a technique to obtain effective representations. In essence, it involves pulling together an `anchor point' and its related samples  while pushing it away from  unrelated samples in the embedding space. Originally developed in self-supervised learning \cite{chen2020simple,henaff2020data}, it has since been adopted in supervised settings \cite{khosla2020supervised} where samples with the same/different labels are deemed related/unrelated with respect to an anchor.

In this work, we use a hierarchical contrastive loss \cite{liang2022zero} alongside the standard binary cross entropy loss on the task outcome probability. On the higher level, this is intended to form distinctive clusters of article-specific case representations. We hypothesize that this distinctiveness maximization constraint in turn helps the article-specific representation component to extract salient information with respect to each article more effectively. At the lower level, inside the latent space for each article, we further perform contrastive learning among cases based on their outcome for tasks A and B, respectively. This allows the positive outcome representations of cases under a specific article to stay closer and separate from the negative outcome cases, leading to formation of sub-clusters. We apply contrastive losses for Task A and Task B separately on the interaction-aware representations.  It is calculated as the mean loss computed based on every article based representation as an anchor point.
The loss for the interaction-aware representation of the $j^\text{th}$ case specific to the $i^\text{th}$ article for Task B $c^\prime_{bi_{j}}$  as an anchor point is calculated as follows:
\begin{equation}
\scalemath{0.7}{
\begin{aligned}
     l(c_{bi_j}^\prime)  =  & \text{log}(\sum_{f \in Q(i,j)} \frac{ \text{exp}(c^\prime_{bi_j} \cdot f /\tau_a)}{\sum_{g \in P(i,j)} \text{exp}( c^\prime_{bi_j}\cdot g /\tau_a)}) \\ & \times \alpha \sum_{f \in R(i,j)} \frac{ \text{exp}(c^\prime_{bi_j} \cdot f /\tau_c)}{\sum_{g \in Q(i,j)} \text{exp}( c^\prime_{bi_j}\cdot g /\tau_c)})
    \end{aligned}}
    \label{contrast}
\end{equation}
where $P(i,j) = \{c^\prime_{bl_m}| l \neq i ~~\&~~  m \neq j \}$ (all representations except the anchor point),  $Q(i,j) = \{c^\prime_{bl_m}| l = i ~~\&~~ m \neq j) \}$ (all representations which share the same article as the anchor point),  $R(i,j) = \{c^\prime_{bl_m}| l = i ~~\&~~ y_{i_j} = y_{l_m} ~~\&~~ m \neq j\}$, $y_{i_j}$ the denotes binary outcome label of case $j$ with respect to article $i$ (all representations which share the same article and outcome of that article as the anchor point). $\tau_a$ and $\tau_c$  are scalar temperature parameters that control the penalties on negative samples. The first term in the above equation \ref{contrast} denotes the contrastive loss among article representations (i.e., for a given anchor point, positive pairs are obtained by article-aware representations of cases in the batch which share the same article with the anchor point; negative pairs are the remaining article representations). The second term contrasts the cases based on their task outcome within specific article representations.  
Contrastive learning has shown to be more effective with larger batch sizes \cite{radford2021learning, chen2020simple}. To account for smaller batch size due to computational constraints incurred with our hierarchical setup, we use a memory bank \cite{wu2018unsupervised} which progressively reuses the representations from previous batches in computing the contrastive loss.

\section{Experiments \& Discussion}
\subsection{Dataset \& Baseline Models}
We experiment on the ECHR task A and B datasets of LexGLUE \cite{chalkidis2022lexglue}, which consist of 11k case fact descriptions chronologically split into training (2001–2016, 9k cases), validation (2016–2017, 1k cases), and test sets (2017-2019, 1k cases). The label set for both tasks include 10 prominent ECHR articles. \noindent Implementation details are given in Appendix \ref{sec:appendix}.  We compare our models against the following baselines:

\noindent \textbf{Task A/B only:} We train single-task models only on task A and B labels, respectively. Their architecture is based on our model with contrastive loss, task dependency and article-specific representation components removed.

\noindent \textbf{Multi-task:} Using the same architecture, we also develop a model for both task A and B that is trained in classic multi-task fashion with separate classification heads without article-specific representations and task dependency components. 

\subsection{Performance Evaluation}
Following \citealt{chalkidis2022lexglue}, we report micro-F1 ($\mu$-F1) and macro-F1 (m-F1) scores across the 10 ECtHR articles contained in the dataset. We also report hard-macro-F1 (hm-F1) for Task A following \citealt{santosh2022deconfounding}, which is the mean F1-score computed for each article where cases with that article having been violated are considered as positive instances, and cases with that article being alleged but not found to have been violated as negative instances. This forms the most critical measure for violation classification as it conditions on allegation.

\noindent We compare the performance of our method with the baselines in the first four rows of Table \ref{tab}. The multi-task method performs better than the individual tasks alone, validating the dependency between them. Our method performs better than the multi-task architecture, highlighting the effectiveness of our feature and label dependency components, as well as of the hierarchical contrastive loss. It scores higher by a small margin in Task B and Task A micro-F1 but achieves a larger benefit (>5\%) in task A macro-F1 and hard macro-F1 metrics. This suggests that our model is of particular utility for sparser classes.  

\begin{table}
\centering
\caption{Results on Task A and B. `feat.', `lab,'. `Cont.' indicate feature, label and contrastive, respectively}
\scalebox{0.7}{
\centering
    \begin{tabular}{|p{3cm}||p{0.9cm}|p{0.9cm}||p{0.9cm}|p{0.9cm}|p{1.1cm}|}
  \hline
   &  \multicolumn{2}{c||}{\textbf{Task B}} & \multicolumn{3}{c|}{\textbf{Task A}}\\
  \hline
   \textbf{Model} & \textbf{$\mu$-F1} &\textbf{m-F1} & \textbf{$\mu$-F1} &\textbf{m-F1} & \textbf{hm-F1}\\ 
 \hline
 Task B only &  76.20&  67.15  & - & - & - \\
 Task A only & - & - &  68.42 & 56.26 & 54.14\\
 Multi-task &  78.17 & 69.16 & 69.29 & 58.05 & 55.57 \\
 Our Method &  79.29  & 70.97 & 71.26 & 65.24 & 60.90 \\
 \hline
 w/o feat. \& lab. & 78.93 & 71.45 & 70.07 &   59.14 &   57.09 \\
 w/o feat. & 78.59 & 71.56 & 70.68 & 63.93 & 59.28 \\
 w/o label &  79.09 & 71.38 &  70.32 & 64.12 & 61.70 \\
 \hline
   only gold lab.  & 78.21 & 70.03  & 81.46 & 78.93 & 66.59 \\
 gold lab. + feat. & 77.68 & 70.40 & 83.19 &  78.79 & 67.42 \\
 \hline
 w/o outcome Cont. & 78.42 & 69.48 &  69.86 & 60.84  & 57.62 \\
 w/o article Cont. & 79.02 & 71.14 &  71.16 & 64.68  & 59.86 \\
 \hline  
 Task A Cont. & - & - &  70.16 & 62.14  & 58.12 \\
 Task B Cont. &  78.16 & 69.42 &- & - & -\\
  \hline
  \end{tabular}}
  \label{tab}
  \vspace{-4mm}
\end{table}

\subsection{Analysis}
\subsubsection{Ablation on Task Dependency Learning}
We create three ablation conditions by removing one interaction mechanism in each of them: 
(i) \emph{w/o feature \& label} (article-specific representation but with no concatenation with features or labels) (ii) \emph{w/o feature} (article-specific representation concatenated with task B label as classified by the model) and (iii) \emph{w/o label} (article-specific representation concatenated with task B features).

\noindent From the second section of Table \ref{tab}, we observe that even the performance of \emph{w/o feature and label} is better than \emph{multi-task} model indicated by a small but consistent margin, indicating that article-specific case representation is a competitive component. Both \emph{w/o feature}, \emph{w/o label} models perform better than \emph{w/o feature \& label} in Task A, demonstrating the benefit of passing on task B model information explicitly. Between them, \emph{w/o label} performs better in m-F1 and hm-F1 scores of Task A than \emph{w/o feature} showing that providing the fine-grained representation of features from the allegation identification model is more useful than the predicted allegation labels only.

Further, to evaluate the impact of allegation identification performance on downstream violation classification, we conduct a control experiment in which we provide the actual gold allegation set of articles as input to the violation component in place of the predicted ones. We create two variants: \emph{only gold labels} and \emph{gold labels + features} (gold labels along with task B features). From the third section of Table \ref{tab}, we observe that performance on Task A increases substantially in micro-F1 and macro-F1, which is intuitive as the model gets access to perfect allegation information. Interestingly, when adding task allegation feature information, we notice an additional small increase in hard macro-F1, indicating that adding feature information can have beneficial effect even in  the presence of gold task B labels. These results form an upper bound of the benefit of using task dependency information from allegation branch to violation branch in our architecture, and the size of the performance gap motivates future work on accurate allegation classification.

\subsubsection{Ablation Hierarchical Contrastive Loss}
We carry out an ablation experiment for each component of our hierarchical contrastive loss: (i) disable article-level contrastive learning to obtain distinctive article representations (`w/o article contrastive') and  (ii) disable outcome-based contrastive learning to separate cases based on outcome within each article cluster (`w/o outcome contrastive'). Intuitively, from the fourth section of Table \ref{tab}, we observe that removing the outcome contrastive loss has the larger effect on performance, as it directly relates to the predicted label. The smaller but consistent drop in performance when removing article-level contrastive learning supports our hypothesis that maximizing the distinctiveness among article representations encourages the model to learn how to extract article-specific salient information from case facts. 

Finally, to study the impact of our contrastive loss component alone, we evaluate single-task models applying our hierarchical contrastive loss referred to as `Task A/B Contrastive'. From the last section of Table \ref{tab}, we  observe that they outperform their baseline counterparts in both tasks, but still stay behind our model.

\section{Conclusion}
We improve ECtHR article violation classification from fact statements by leveraging feature and label information from an allegation classification model. We also leveraged hierarchical contrastive loss to contrast between different article representations and case representations based on outcome with respect to a specific article. Given a static pre-trained encoder, our models outperform a straightforward multi-task architecture by a small but consistent margin, with larger improvements for sparse classes. These results suggest that the tasks of allegation and violation classification on ECtHR fact statements interrelate in a way that may not be optimally captured using straightforward multi-task architectures, and motivate further research on dependency modeling between related legal classification tasks.

\section*{Limitations}
In this work, we have demonstrated improvements in violation classification for ECtHR cases by leveraging the dependency from allegation classification at the feature and label level. While our feature transfer and contrastive learning techniques are general, our experimental contributions are contextual to the court. The nature of the fact statements as being influenced by the eventual case outcome and not suitable for prospective prediction has already been addressed in section \ref{intro}.

While similar tasks (i.e., allegations paired with findings) exist in many other jurisdictions, they will differ in, for example, legal nature, semantic difficulty, and sub-task dependency. In our case, we choose the allegation classification as the auxiliary task which is closely related and also forms a necessary sub-part of deriving the final outcome related to the main task (violation prediction) at hand. We leave an exploration of the relatedness of auxiliary tasks and their impact on LJP/COC for future work.

One major challenge is dealing with long input case facts description, which is currently handled with hierarchical model in this work. These hierarchical models do not allow tokens in one sentence to attend to tokens in other sentence which leads to sub-optimal interaction modelling. This modeling impact on performance is still underexplored except some preliminary empirical work in  \citealt{dai2022revisiting,chalkidis2022exploration}. Additionally, we freeze the weights in the LegalBERT encoder, both to save computational resources and to evaluate the effectiveness of our dependency mechanism in a clean way.

\section*{Ethics Statement}
We employ pre-trained language models and do not train them from scratch, thus inheriting the biases they may have acquired from their training corpus. Our experiments have been carried out on a dataset of ECtHR decisions which is publicly available as the part of LexGLUE benchmark \cite{chalkidis2022lexglue} and has been derived from the public court database HUDOC\footnote{\url{https://hudoc.echr.coe.int}}. Though these decisions are not anonymized and contain the real names of the involved parties, we do not foresee any harm incurred by our experiments beyond making this information available. This collection of decision documents  is of course historical data and using it to train model may result in classifiers that exhibit behavior that may be considered biased. For example, \citealt{chalkidis2022fairlex} explores disparities in classification performance with regard to an applicant's gender, their age, and the identity of the respondent state. If COC models are deployed as part of a decision support systems, then they of course must be screened for performance/error differences in between groups that are to be treated equally. 

The task of LJP/COC in itself raises serious ethical and legal concerns, both in general and specific to the European Court of Human Rights. However, we do not advocate for the practical adoption of LJP/COC systems by courts. Our prior work in \citealt{santosh2022deconfounding} demonstrates that these systems rely on several shallow surface-level spurious signals that are statistically predictive but legally irrelevant.  This highlights the risk of using predictive systems in high stakes domains such as law. In the same work, we argue that models leveraging the case outcome signal for analytical purposes must be developed mindfully and with the goal of aligning their inferences with legal expert reasoning. This further parallels the broader legal NLP community increasingly addressing ethical aspects of developed systems in the context of technical research (e.g., \citealt{wang2021equality, medvedeva2021automatic, medvedeva2022rethinking, tsarapatsanis2021ethical, leins2020give}). 

In this work, we use COC as a technical benchmarking task that allows the development and study of neural NLP models on legal text. We focus on how to leverage dependencies on two successive tasks (allegation identification and violation classification) based on case facts, as well as on learning effective representations of these facts using contrastive learning. Our results are hence to be understood as technical contributions in pursuit of the overarching goal of developing models capable of deriving insight from data that can be used legally, ethically, and mindfully by experts in solving problems arising in legal research and practice.

All experiments were carried out using Google Colab. We did not track computation hours. 

\bibliography{anthology,custom}

\begin{thebibliography}{48}
\expandafter\ifx\csname natexlab\endcsname\relax\def\natexlab#1{#1}\fi

\bibitem[{Aletras et~al.(2016)Aletras, Tsarapatsanis, Preo{\c{t}}iuc-Pietro,
  and Lampos}]{aletras2016predicting}
Nikolaos Aletras, Dimitrios Tsarapatsanis, Daniel Preo{\c{t}}iuc-Pietro, and
  Vasileios Lampos. 2016.
\newblock Predicting judicial decisions of the european court of human rights:
  A natural language processing perspective.
\newblock \emph{PeerJ Computer Science}, 2:e93.

\bibitem[{Bertalan and Ruiz(2020)}]{bertalan2020predicting}
Vithor Gomes~Ferreira Bertalan and Evandro Eduardo~Seron Ruiz. 2020.
\newblock Predicting judicial outcomes in the brazilian legal system using
  textual features.
\newblock In \emph{DHandNLP@ PROPOR}, pages 22--32.

\bibitem[{Chalkidis et~al.(2019)Chalkidis, Androutsopoulos, and
  Aletras}]{chalkidis2019neural}
Ilias Chalkidis, Ion Androutsopoulos, and Nikolaos Aletras. 2019.
\newblock Neural legal judgment prediction in english.
\newblock In \emph{Proceedings of the 57th Annual Meeting of the Association
  for Computational Linguistics}, pages 4317--4323.

\bibitem[{Chalkidis et~al.(2022{\natexlab{a}})Chalkidis, Dai, Fergadiotis,
  Malakasiotis, and Elliott}]{chalkidis2022exploration}
Ilias Chalkidis, Xiang Dai, Manos Fergadiotis, Prodromos Malakasiotis, and
  Desmond Elliott. 2022{\natexlab{a}}.
\newblock An exploration of hierarchical attention transformers for efficient
  long document classification.
\newblock \emph{arXiv preprint arXiv:2210.05529}.

\bibitem[{Chalkidis et~al.(2020)Chalkidis, Fergadiotis, Malakasiotis, Aletras,
  and Androutsopoulos}]{chalkidis2020legal}
Ilias Chalkidis, Manos Fergadiotis, Prodromos Malakasiotis, Nikolaos Aletras,
  and Ion Androutsopoulos. 2020.
\newblock Legal-bert: The muppets straight out of law school.
\newblock In \emph{Findings of the Association for Computational Linguistics:
  EMNLP 2020}, pages 2898--2904.

\bibitem[{Chalkidis et~al.(2021)Chalkidis, Fergadiotis, Tsarapatsanis, Aletras,
  Androutsopoulos, and Malakasiotis}]{chalkidis2021paragraph}
Ilias Chalkidis, Manos Fergadiotis, Dimitrios Tsarapatsanis, Nikolaos Aletras,
  Ion Androutsopoulos, and Prodromos Malakasiotis. 2021.
\newblock Paragraph-level rationale extraction through regularization: A case
  study on european court of human rights cases.
\newblock In \emph{Proceedings of the 2021 Conference of the North American
  Chapter of the Association for Computational Linguistics: Human Language
  Technologies}, pages 226--241.

\bibitem[{Chalkidis et~al.(2022{\natexlab{b}})Chalkidis, Jana, Hartung,
  Bommarito, Androutsopoulos, Katz, and Aletras}]{chalkidis2022lexglue}
Ilias Chalkidis, Abhik Jana, Dirk Hartung, Michael Bommarito, Ion
  Androutsopoulos, Daniel Katz, and Nikolaos Aletras. 2022{\natexlab{b}}.
\newblock Lexglue: A benchmark dataset for legal language understanding in
  english.
\newblock In \emph{Proceedings of the 60th Annual Meeting of the Association
  for Computational Linguistics (Volume 1: Long Papers)}, pages 4310--4330.

\bibitem[{Chalkidis et~al.(2022{\natexlab{c}})Chalkidis, Pasini, Zhang, Tomada,
  Schwemer, and S{\o}gaard}]{chalkidis2022fairlex}
Ilias Chalkidis, Tommaso Pasini, Sheng Zhang, Letizia Tomada, Sebastian
  Schwemer, and Anders S{\o}gaard. 2022{\natexlab{c}}.
\newblock Fairlex: A multilingual benchmark for evaluating fairness in legal
  text processing.
\newblock In \emph{Proceedings of the 60th Annual Meeting of the Association
  for Computational Linguistics (Volume 1: Long Papers)}, pages 4389--4406.

\bibitem[{Chen et~al.(2020)Chen, Kornblith, Norouzi, and
  Hinton}]{chen2020simple}
Ting Chen, Simon Kornblith, Mohammad Norouzi, and Geoffrey Hinton. 2020.
\newblock A simple framework for contrastive learning of visual
  representations.
\newblock In \emph{International conference on machine learning}, pages
  1597--1607. PMLR.

\bibitem[{Clavi{\'e} and Alphonsus(2021)}]{clavie2021unreasonable}
Benjamin Clavi{\'e} and Marc Alphonsus. 2021.
\newblock The unreasonable effectiveness of the baseline: Discussing svms in
  legal text classification.
\newblock In \emph{Legal Knowledge and Information Systems}, pages 58--61. IOS
  Press.

\bibitem[{Dai et~al.(2022)Dai, Chalkidis, Darkner, and
  Elliott}]{dai2022revisiting}
Xiang Dai, Ilias Chalkidis, Sune Darkner, and Desmond Elliott. 2022.
\newblock \href {https://aclanthology.org/2022.findings-emnlp.534} {Revisiting
  transformer-based models for long document classification}.
\newblock In \emph{Findings of the Association for Computational Linguistics:
  EMNLP 2022}, pages 7212--7230, Abu Dhabi, United Arab Emirates. Association
  for Computational Linguistics.

\bibitem[{Henaff(2020)}]{henaff2020data}
Olivier Henaff. 2020.
\newblock Data-efficient image recognition with contrastive predictive coding.
\newblock In \emph{International conference on machine learning}, pages
  4182--4192. PMLR.

\bibitem[{Katz et~al.(2017)Katz, Bommarito, and Blackman}]{katz2017general}
Daniel~Martin Katz, Michael~J Bommarito, and Josh Blackman. 2017.
\newblock A general approach for predicting the behavior of the supreme court
  of the united states.
\newblock \emph{PloS one}, 12(4):e0174698.

\bibitem[{Kaufman et~al.(2019)Kaufman, Kraft, and Sen}]{kaufman2019improving}
Aaron~Russell Kaufman, Peter Kraft, and Maya Sen. 2019.
\newblock Improving supreme court forecasting using boosted decision trees.
\newblock \emph{Political Analysis}, 27(3):381--387.

\bibitem[{Khosla et~al.(2020)Khosla, Teterwak, Wang, Sarna, Tian, Isola,
  Maschinot, Liu, and Krishnan}]{khosla2020supervised}
Prannay Khosla, Piotr Teterwak, Chen Wang, Aaron Sarna, Yonglong Tian, Phillip
  Isola, Aaron Maschinot, Ce~Liu, and Dilip Krishnan. 2020.
\newblock Supervised contrastive learning.
\newblock \emph{Advances in Neural Information Processing Systems},
  33:18661--18673.

\bibitem[{Kingma and Ba(2015)}]{kingma2014adam}
Diederik~P Kingma and Jimmy Ba. 2015.
\newblock Adam: a method for stochastic optimization 3rd int.
\newblock In \emph{International Conference on Learning Representations}.

\bibitem[{Kowsrihawat et~al.(2018)Kowsrihawat, Vateekul, and
  Boonkwan}]{kowsrihawat2018predicting}
Kankawin Kowsrihawat, Peerapon Vateekul, and Prachya Boonkwan. 2018.
\newblock Predicting judicial decisions of criminal cases from thai supreme
  court using bi-directional gru with attention mechanism.
\newblock In \emph{2018 5th Asian Conference on Defense Technology (ACDT)},
  pages 50--55. IEEE.

\bibitem[{Leins et~al.(2020)Leins, Lau, and Baldwin}]{leins2020give}
Kobi Leins, Jey~Han Lau, and Timothy Baldwin. 2020.
\newblock Give me convenience and give her death: Who should decide what uses
  of nlp are appropriate, and on what basis?
\newblock In \emph{Proceedings of the 58th Annual Meeting of the Association
  for Computational Linguistics}, pages 2908--2913.

\bibitem[{Liang et~al.(2022)Liang, Chen, Gui, He, Yang, and Xu}]{liang2022zero}
Bin Liang, Zixiao Chen, Lin Gui, Yulan He, Min Yang, and Ruifeng Xu. 2022.
\newblock Zero-shot stance detection via contrastive learning.
\newblock In \emph{Proceedings of the ACM Web Conference 2022}, pages
  2738--2747.

\bibitem[{Luo et~al.(2017)Luo, Feng, Xu, Zhang, and Zhao}]{luo2017learning}
Bingfeng Luo, Yansong Feng, Jianbo Xu, Xiang Zhang, and Dongyan Zhao. 2017.
\newblock Learning to predict charges for criminal cases with legal basis.
\newblock In \emph{Proceedings of the 2017 Conference on Empirical Methods in
  Natural Language Processing}, pages 2727--2736.

\bibitem[{Malik et~al.(2021)Malik, Sanjay, Nigam, Ghosh, Guha, Bhattacharya,
  and Modi}]{malik2021ildc}
Vijit Malik, Rishabh Sanjay, Shubham~Kumar Nigam, Kripabandhu Ghosh,
  Shouvik~Kumar Guha, Arnab Bhattacharya, and Ashutosh Modi. 2021.
\newblock Ildc for cjpe: Indian legal documents corpus for court judgment
  prediction and explanation.
\newblock In \emph{Proceedings of the 59th Annual Meeting of the Association
  for Computational Linguistics and the 11th International Joint Conference on
  Natural Language Processing (Volume 1: Long Papers)}, pages 4046--4062.

\bibitem[{Medvedeva et~al.(2021)Medvedeva, {\"U}st{\"u}n, Xu, Vols, and
  Wieling}]{medvedeva2021automatic}
Masha Medvedeva, Ahmet {\"U}st{\"u}n, Xiao Xu, Michel Vols, and Martijn
  Wieling. 2021.
\newblock Automatic judgement forecasting for pending applications of the
  european court of human rights.
\newblock In \emph{ASAIL/LegalAIIA@ ICAIL}.

\bibitem[{Medvedeva et~al.(2020)Medvedeva, Vols, and
  Wieling}]{medvedeva2020using}
Masha Medvedeva, Michel Vols, and Martijn Wieling. 2020.
\newblock Using machine learning to predict decisions of the european court of
  human rights.
\newblock \emph{Artificial Intelligence and Law}, 28(2):237--266.

\bibitem[{Medvedeva et~al.(2022)Medvedeva, Wieling, and
  Vols}]{medvedeva2022rethinking}
Masha Medvedeva, Martijn Wieling, and Michel Vols. 2022.
\newblock Rethinking the field of automatic prediction of court decisions.
\newblock \emph{Artificial Intelligence and Law}, pages 1--18.

\bibitem[{Mumcuo{\u{g}}lu et~al.(2021)Mumcuo{\u{g}}lu, {\"O}zt{\"u}rk, Ozaktas,
  and Ko{\c{c}}}]{mumcuouglu2021natural}
Emre Mumcuo{\u{g}}lu, Ceyhun~E {\"O}zt{\"u}rk, Haldun~M Ozaktas, and Aykut
  Ko{\c{c}}. 2021.
\newblock Natural language processing in law: Prediction of outcomes in the
  higher courts of turkey.
\newblock \emph{Information Processing \& Management}, 58(5):102684.

\bibitem[{Niklaus et~al.(2021)Niklaus, Chalkidis, and
  St{\"u}rmer}]{niklaus2021swiss}
Joel Niklaus, Ilias Chalkidis, and Matthias St{\"u}rmer. 2021.
\newblock Swiss-judgment-prediction: A multilingual legal judgment prediction
  benchmark.
\newblock In \emph{Proceedings of the Natural Legal Language Processing
  Workshop 2021}, pages 19--35.

\bibitem[{Radford et~al.(2021)Radford, Kim, Hallacy, Ramesh, Goh, Agarwal,
  Sastry, Askell, Mishkin, Clark et~al.}]{radford2021learning}
Alec Radford, Jong~Wook Kim, Chris Hallacy, Aditya Ramesh, Gabriel Goh,
  Sandhini Agarwal, Girish Sastry, Amanda Askell, Pamela Mishkin, Jack Clark,
  et~al. 2021.
\newblock Learning transferable visual models from natural language
  supervision.
\newblock In \emph{International Conference on Machine Learning}, pages
  8748--8763. PMLR.

\bibitem[{Santosh et~al.(2023)Santosh, Ichim, and Grabmair}]{santosh2023Zero}
T.~Y. S.~S Santosh, Oana Ichim, and Matthias Grabmair. 2023.
\newblock Zero shot transfer of article-aware legal outcome classification for
  european court of human rights cases.
\newblock \emph{arXiv preprint arXiv:2302.00609}.

\bibitem[{Santosh et~al.(2022)Santosh, Xu, Ichim, and
  Grabmair}]{santosh2022deconfounding}
T.y.s.s Santosh, Shanshan Xu, Oana Ichim, and Matthias Grabmair. 2022.
\newblock \href {https://aclanthology.org/2022.emnlp-main.74} {Deconfounding
  legal judgment prediction for {E}uropean court of human rights cases towards
  better alignment with experts}.
\newblock In \emph{Proceedings of the 2022 Conference on Empirical Methods in
  Natural Language Processing}, pages 1120--1138, Abu Dhabi, United Arab
  Emirates. Association for Computational Linguistics.

\bibitem[{Sert et~al.(2021)Sert, Y{\i}ld{\i}r{\i}m, and
  Ha{\c{s}}lak}]{sert2021using}
Mehmet~Fatih Sert, Engin Y{\i}ld{\i}r{\i}m, and {\.I}rfan Ha{\c{s}}lak. 2021.
\newblock Using artificial intelligence to predict decisions of the turkish
  constitutional court.
\newblock \emph{Social Science Computer Review}, page 08944393211010398.

\bibitem[{Shaikh et~al.(2020)Shaikh, Sahu, and Anand}]{shaikh2020predicting}
Rafe~Athar Shaikh, Tirath~Prasad Sahu, and Veena Anand. 2020.
\newblock Predicting outcomes of legal cases based on legal factors using
  classifiers.
\newblock \emph{Procedia Computer Science}, 167:2393--2402.

\bibitem[{Srivastava et~al.(2014)Srivastava, Hinton, Krizhevsky, Sutskever, and
  Salakhutdinov}]{srivastava2014dropout}
Nitish Srivastava, Geoffrey Hinton, Alex Krizhevsky, Ilya Sutskever, and Ruslan
  Salakhutdinov. 2014.
\newblock Dropout: a simple way to prevent neural networks from overfitting.
\newblock \emph{The journal of machine learning research}, 15(1):1929--1958.

\bibitem[{Strickson and De~La~Iglesia(2020)}]{strickson2020legal}
Benjamin Strickson and Beatriz De~La~Iglesia. 2020.
\newblock Legal judgement prediction for uk courts.
\newblock In \emph{Proceedings of the 2020 the 3rd international conference on
  information science and system}, pages 204--209.

\bibitem[{{\c{S}}ulea et~al.(2017){\c{S}}ulea, Zampieri, Vela, and van
  Genabith}]{csulea2017predicting}
Octavia-Maria {\c{S}}ulea, Marcos Zampieri, Mihaela Vela, and Josef van
  Genabith. 2017.
\newblock Predicting the law area and decisions of french supreme court cases.
\newblock In \emph{Proceedings of the International Conference Recent Advances
  in Natural Language Processing, RANLP 2017}, pages 716--722.

\bibitem[{Tsarapatsanis and Aletras(2021)}]{tsarapatsanis2021ethical}
Dimitrios Tsarapatsanis and Nikolaos Aletras. 2021.
\newblock On the ethical limits of natural language processing on legal text.
\newblock In \emph{Findings of the Association for Computational Linguistics:
  ACL-IJCNLP 2021}, pages 3590--3599.

\bibitem[{Valvoda et~al.(2023)Valvoda, Cotterell, and Teufel}]{valvoda2022role}
Josef Valvoda, Ryan Cotterell, and Simone Teufel. 2023.
\newblock On the role of negative precedent in legal outcome prediction.
\newblock \emph{Transactions of the Association for Computational Linguistics},
  11:34--48.

\bibitem[{Vaswani et~al.(2017)Vaswani, Shazeer, Parmar, Uszkoreit, Jones,
  Gomez, Kaiser, and Polosukhin}]{vaswani2017attention}
Ashish Vaswani, Noam Shazeer, Niki Parmar, Jakob Uszkoreit, Llion Jones,
  Aidan~N Gomez, {\L}ukasz Kaiser, and Illia Polosukhin. 2017.
\newblock Attention is all you need.
\newblock \emph{Advances in neural information processing systems}, 30.

\bibitem[{Virtucio et~al.(2018)Virtucio, Aborot, Abonita, Avinante, Copino,
  Neverida, Osiana, Peramo, Syjuco, and Tan}]{virtucio2018predicting}
Michael Benedict~L Virtucio, Jeffrey~A Aborot, John Kevin~C Abonita, Roxanne~S
  Avinante, Rother Jay~B Copino, Michelle~P Neverida, Vanesa~O Osiana, Elmer~C
  Peramo, Joanna~G Syjuco, and Glenn Brian~A Tan. 2018.
\newblock Predicting decisions of the philippine supreme court using natural
  language processing and machine learning.
\newblock In \emph{2018 IEEE 42nd annual computer software and applications
  conference (COMPSAC)}, volume~2, pages 130--135. IEEE.

\bibitem[{Waltl et~al.(2017)Waltl, Bonczek, Scepankova, Landthaler, and
  Matthes}]{waltl2017predicting}
Bernhard Waltl, Georg Bonczek, Elena Scepankova, J{\"o}rg Landthaler, and
  Florian Matthes. 2017.
\newblock Predicting the outcome of appeal decisions in germany’s tax law.
\newblock In \emph{International conference on electronic participation}, pages
  89--99. Springer.

\bibitem[{Wang et~al.(2021)Wang, Xiao, Ma, Zhong, Tu, Zhang, Liu, and
  Sun}]{wang2021equality}
Yuzhong Wang, Chaojun Xiao, Shirong Ma, Haoxi Zhong, Cunchao Tu, Tianyang
  Zhang, Zhiyuan Liu, and Maosong Sun. 2021.
\newblock Equality before the law: Legal judgment consistency analysis for
  fairness.
\newblock \emph{arXiv e-prints}, pages arXiv--2103.

\bibitem[{Wu et~al.(2018)Wu, Xiong, Yu, and Lin}]{wu2018unsupervised}
Zhirong Wu, Yuanjun Xiong, Stella~X Yu, and Dahua Lin. 2018.
\newblock Unsupervised feature learning via non-parametric instance
  discrimination.
\newblock In \emph{Proceedings of the IEEE conference on computer vision and
  pattern recognition}, pages 3733--3742.

\bibitem[{Xu et~al.(2020)Xu, Wang, Chen, Pan, Wang, and
  Zhao}]{xu2020distinguish}
Nuo Xu, Pinghui Wang, Long Chen, Li~Pan, Xiaoyan Wang, and Junzhou Zhao. 2020.
\newblock Distinguish confusing law articles for legal judgment prediction.
\newblock In \emph{Proceedings of the 58th Annual Meeting of the Association
  for Computational Linguistics}, pages 3086--3095.

\bibitem[{Yang et~al.(2019)Yang, Jia, Zhou, and Luo}]{yang2019legal}
Wenmian Yang, Weijia Jia, Xiaojie Zhou, and Yutao Luo. 2019.
\newblock Legal judgment prediction via multi-perspective bi-feedback network.
\newblock In \emph{Proceedings of the 28th International Joint Conference on
  Artificial Intelligence}, pages 4085--4091.

\bibitem[{Yang et~al.(2016)Yang, Yang, Dyer, He, Smola, and
  Hovy}]{yang2016hierarchical}
Zichao Yang, Diyi Yang, Chris Dyer, Xiaodong He, Alex Smola, and Eduard Hovy.
  2016.
\newblock Hierarchical attention networks for document classification.
\newblock In \emph{Proceedings of the 2016 conference of the North American
  chapter of the association for computational linguistics: human language
  technologies}, pages 1480--1489.

\bibitem[{Yue et~al.(2021)Yue, Liu, Jin, Wu, Zhang, An, Cheng, Yin, and
  Wu}]{yue2021neurjudge}
Linan Yue, Qi~Liu, Binbin Jin, Han Wu, Kai Zhang, Yanqing An, Mingyue Cheng,
  Biao Yin, and Dayong Wu. 2021.
\newblock Neurjudge: a circumstance-aware neural framework for legal judgment
  prediction.
\newblock In \emph{Proceedings of the 44th International ACM SIGIR Conference
  on Research and Development in Information Retrieval}, pages 973--982.

\bibitem[{Zhang et~al.(2023)Zhang, Dou, Zhu, and Wen}]{zhang2023contrastive}
Han Zhang, Zhicheng Dou, Yutao Zhu, and Ji-Rong Wen. 2023.
\newblock Contrastive learning for legal judgment prediction.
\newblock \emph{ACM Transactions on Information Systems}.

\bibitem[{Zhong et~al.(2018)Zhong, Guo, Tu, Xiao, Liu, and
  Sun}]{zhong2018legal}
Haoxi Zhong, Zhipeng Guo, Cunchao Tu, Chaojun Xiao, Zhiyuan Liu, and Maosong
  Sun. 2018.
\newblock Legal judgment prediction via topological learning.
\newblock In \emph{Proceedings of the 2018 Conference on Empirical Methods in
  Natural Language Processing}, pages 3540--3549.

\bibitem[{Zhong et~al.(2020)Zhong, Wang, Tu, Zhang, Liu, and
  Sun}]{zhong2020iteratively}
Haoxi Zhong, Yuzhong Wang, Cunchao Tu, Tianyang Zhang, Zhiyuan Liu, and Maosong
  Sun. 2020.
\newblock Iteratively questioning and answering for interpretable legal
  judgment prediction.
\newblock In \emph{Proceedings of the AAAI Conference on Artificial
  Intelligence}, volume~34, pages 1250--1257.

\end{thebibliography}
\bibliographystyle{acl_natbib}

\appendix

\section{Implementation Details}
\label{sec:appendix}
Our models compute word embeddings of size 768. Our word level attention context vector size is 300. The sentence level GRU encoder dimension is 200, thus giving a bidirectional embedding of size 400, and a sentence level attention vector dimension of 200. The final dense classifier for all tasks has 100 hidden units. We use a mini batches size of 32 and the model is optimized end-to-end using Adam  \cite{kingma2014adam}. The dropout rate \cite{srivastava2014dropout} in all layers is 0.1. We determine the best learning rate using grid search on the development set and use early stopping based on the development set m-F1 score. We finetuned   $\tau_a$, $\tau_c$ with an additional constraint of $\tau_a$ < $\tau_c$  among the values of $\{0.07,0.1,0.14,0.2, 0.25, 0.3\}$ so that it aids in pulling together the representations belonging to the same article in latent space (leading to distinct article clusters) and also in further slightly pulling together the representations of cases belonging to the same outcome in each separated article-specific embedding latent space compared to the other outcome cases in that same article. We set $\alpha$ to be 0.5. We use a memory bank of size 32 per article and outcome, and store only the most recent examples per article and its corresponding outcome.


\end{document}